# Covariate conscious approach for Gait recognition based upon Zernike moment invariants

Himanshu Aggarwal, *Student Member, IEEE* & Dinesh K. Vishwakarma, *Member, IEEE*

*Abstract*— Gait recognition i.e. identification of an individual from his/her walking pattern is an emerging field. While existing gait recognition techniques perform satisfactorily in normal walking conditions, there performance tend to suffer drastically with variations in clothing and carrying conditions. In this work, we propose a novel covariate cognizant framework to deal with the presence of such covariates. We describe gait motion by forming a single 2D spatio-temporal template from video sequence, called Average Energy Silhouette image (AESI). Zernike moment invariants (ZMIs) are then computed to screen the parts of AESI infected with covariates. Following this, features are extracted from Spatial Distribution of Oriented Gradients (SDOGs) and novel Mean of Directional Pixels (MDPs) methods. The obtained features are fused together to form the final well-endowed feature set. Experimental evaluation of the proposed framework on three publicly available datasets i.e. CASIA dataset B, OU-ISIR Treadmill dataset B and USF Human-ID challenge dataset with recently published gait recognition approaches, prove its superior performance.

## I. INTRODUCTION

BIOMETRIC have become a common sight these days. While fingerprint scanner, iris scanner, palm scanner etc. are being extensively used in offices, banks, legal organizations; yet there are some biometric methods whose usage in daily lives has still been much constricted.

Individual recognition on the basis of gait is one such domain. The domain exploits the notion that every individual has its own idiosyncratic way of walking. Also, gait data collection being non-intrusive, contact-free and taking into account its easy availability, owing to the widespread usage of surveillance cameras, make gait biometrics a very promising field. However the efficiency of gait recognition algorithms do suffer from presence of some external factors such as covariates like clothing, footwear, bags, etc. This makes gait biometrics a challenging task but nevertheless useful.

Gait recognition approaches can be classified into broadly two categories: Model based approach and Model-free (Appearance based) approach. Model based approach [1] [2] [3] [4] [5] employ modelling of human body structure and local movement patterns of different body parts. The parameters of the models, like knee position, pelvis to knee distance, hip position etc. are mostly learnt by processing the gait sequence. Model based approach is generally view and scale invariant. However, the performance of this approach is highly dependent on the quality of extracted human silhouettes, low quality of which can lead to inaccurate and incorrect parameter estimation resulting in plummeting of the performance. On the other hand, Model free approach [6] [7] [8] [9] [10]is not explicitly based on structural information. This approach works on temporal and shape information. Compared to model-based approach, model-free approach is computationally less expensive and use of temporal information results in much better recognition performance. Furthermore, model-free approach is more robust to noise and has thus become the preferred approach in research.

One of the primary challenge for vision based gait recognition techniques is to disentangle the identity-unrelated factors which yet alter gait appearances drastically. Though existing works [6] [7] [10] [11] [12] [13], give satisfactory performance in normal walking conditions taken under controlled environment, these methods are sensitive to silhouette distortion which occur due to the presence of covariates. A representative gallery set is required for such metrics to perform effectively. In real-world scenarios such an assumption does not hold true. Since the conditions in which a query gait sequence is collected is mostly unknown, a gallery set encapsulating entire population cannot be obtained.

To address these problems, a unified covariate cognizant approach to gait recognition has been proposed in this paper. The method works on spatio-temporal image representation of gait named AESI [14]. AESI is a compact representation of gait which preserves the necessary shape and temporal information of the gait sequence. Furthermore, the presented framework employ two stages. The first stage involves detection of anomalies due to presence of covariates. Zernike invariant moments [15] are used for this purpose. The moments are computed for the AESI and then the distance

Himanshu Aggarwal is working as an Associate Software Engineer in the Division of Computer Vision at Qualcomm Incorporated, Hyderabad, India (e-mail:hiaggarw@qti.qualcomm.com, himanshub43@gmail.com)
D.K. Vishwakarma is working as an Assistant Professor in the Department of Electronics and Communication Engineering, Delhi Technological University, Delhi, India-11042 (e-mail: dinesh@dtu.ac.in, dvishwakarma@gmail.com).

between the moments is used as a comparison metric. Thereafter, second stage involve feature extraction from parts of AESI devoid of any anomalies as decided by the comparison in the first stage. For this, Spatial distribution of gradients (SDOG) [14] is employed. SDOG calculates local orientation of gradients over the silhouette image (AESI) and its sub-pyramids creating feature set that encapsulates both local and global information about the image. In addition to SDOG, MDPs are computed and integrated in the feature set. MDPs show distribution of the MDPs in $x$ direction, the direction in which walking motion is most prominent [16]. MDPs are computed only in regions of high temporality, typically the limb and neck area, and preserve structural information in these regions. As a result the final formed feature set has extensive distinguishing capability, in addition to being low dimensional, resulting in optimal recognition performance. Moreover, the classical approach employed in existing works [6] [7] [17] is to use Principal component analysis(PCA) and Multiple Discriminant Analysis(MDA) approach for feature set dimension reduction obtained from direct matching on various kinds of spatio temporal representations of gait. Our method intrinsically generate a reliable and compact feature set, thus making the use of these approaches superfluous.

Experimental results on three popular gait datasets demonstrate the efficacy of our approach. Our method achieved an overall recognition accuracy of 72.7% on OU-ISIR Treadmill B dataset [18], this is better than recent GII-SF approach [19] by 11.5%. For CASIA B dataset [20], the improvement is even more, where our method achieved overall accuracy of 91.47% which is nearly 19% better than the GII-SF. We also achieved accuracy of 72.53% on the USF Human-ID challenge dataset [21] which slightly outperforms the very recent clothing and carrying condition invariant rotation forest approach [22]. The key contributions of this work are as follows:

- The study of a novel consolidated approach to gait recognition that is capable of detecting covariates and minimize their repercussions by employing AESI, ZMIs, SDOGs and MDPs.
- Employment of AESIs that effectively represents temporal and shape information by fusing gait silhouette images using reference point alignment and forming a single image template .
- Introduction of a novel method for determination of presence of covariates by segmenting AESI into multiple parts and using ZMIs as a comparison metric.
- Use of SDOGs method for defining gait characteristics, that is based on local gradient orientation for feature extraction from gait template and provides low dimensional feature set that encompass both local and global viable information.
- A novel MDPs method that extracts structural information from high temporal regions by studying variation of MDPs in horizontal direction, the direction in which gait motion is most prominent.

The remainder of the paper is organized as follows: A brief survey of the existing popular gait recognition techniques is done in Section 2. Section 3 discusses the pipeline of the proposed approach and details the various modules of the pipeline. A detailed analysis of the performance of the proposed technique is performed in Section 4, alongside comparisons with existing similar state-of-the-art methods. Section 6 concludes the paper.

## II. RELATED WORK

In recent years, several techniques have been devised for the recognition of individuals through their gait. As already discussed two approaches to gait recognition exist -Model based approach and Appearance based approach. In model based approach human body structure is modelled through various geometrical shapes. Johansson *[1]* introduced Moving light display i.e., MLD, a unique technique for gait recognition. MLD involved fastening of bright markers to human clothing in order to capture gait motion of subject. Ben Abdelkader et al. *[2]* argued that gait stride and cadence can be used for identification of individuals. The method estimated stride length using a calibrated camera and then used Bayesian classifier for identification. Johnson and Bobick *[3]* inspected a method to recognize people using static body parameters like height of individual, distance between head and pelvis, distance between pelvis and left foot, and distance between right and left foot. Liu and Sarkar *[23]* proposed averaging of silhouette frames for determination of gait characteristics of an individual. The method then used Euclidean distance between averaged silhouettes for classification. Lee et al *[24]* proposed Shape Variated-based frieze pattern (SVB-frieze pattern) representation for gait. SVB-frieze pattern was formed by projecting difference frames onto 1 dimension, the projected values of difference frames were then summed over a gait cycle to obtain the frieze pattern. Nixon et al *[4] [25]* presented automatic gait recognition involving estimation of rotation pattern of hips and modelling of leg movement using a pendulum model. K-nearest neighbor classifier was then used for classification.

Model free approaches involve period based methods. These methods provide compact representation of gait by extracting features from a single image obtained from one gait cycle. Chai et al. *[26]* introduced Perceptual Shape Descriptor (PSD) that encompassed shape information of walking silhouettes. The formed PSDs were then accumulated over time to form Perceptual curve, which was used for gait recognition. Tan et al. *[27]* formed frieze pattern by projecting normalized silhouettes along four directions- vertical, horizontal and two diagonals. The resultant feature set was dimensionally reduced using PCA. A Mahalanobis distance based nearest neighbour method was then used for classification. Han and Bhanu introduced concept of Gait energy image (GEI) *[6]*, which involved averaging of binary silhouette images over a gait cycle to form single image. GEI still remains one of the most popular technique for gait recognition and numerous works use GEI as underpinning. Tao et al. *[11]* used Gabor functions over GEI for image understanding and employed a General Tensor Discriminant

Analysis for recognition. A matrix based Marginal Fisher analysis (MFA) was presented by Xu et al. *[10]*, that produced a compact feature set, to address problem of gait recognition.

Gait flow image(GFI) which represented motion information in a gait cycle by optical flow. Recently, Parul et al. proposed Gait Information image (GII) *[19]*, a technique based on

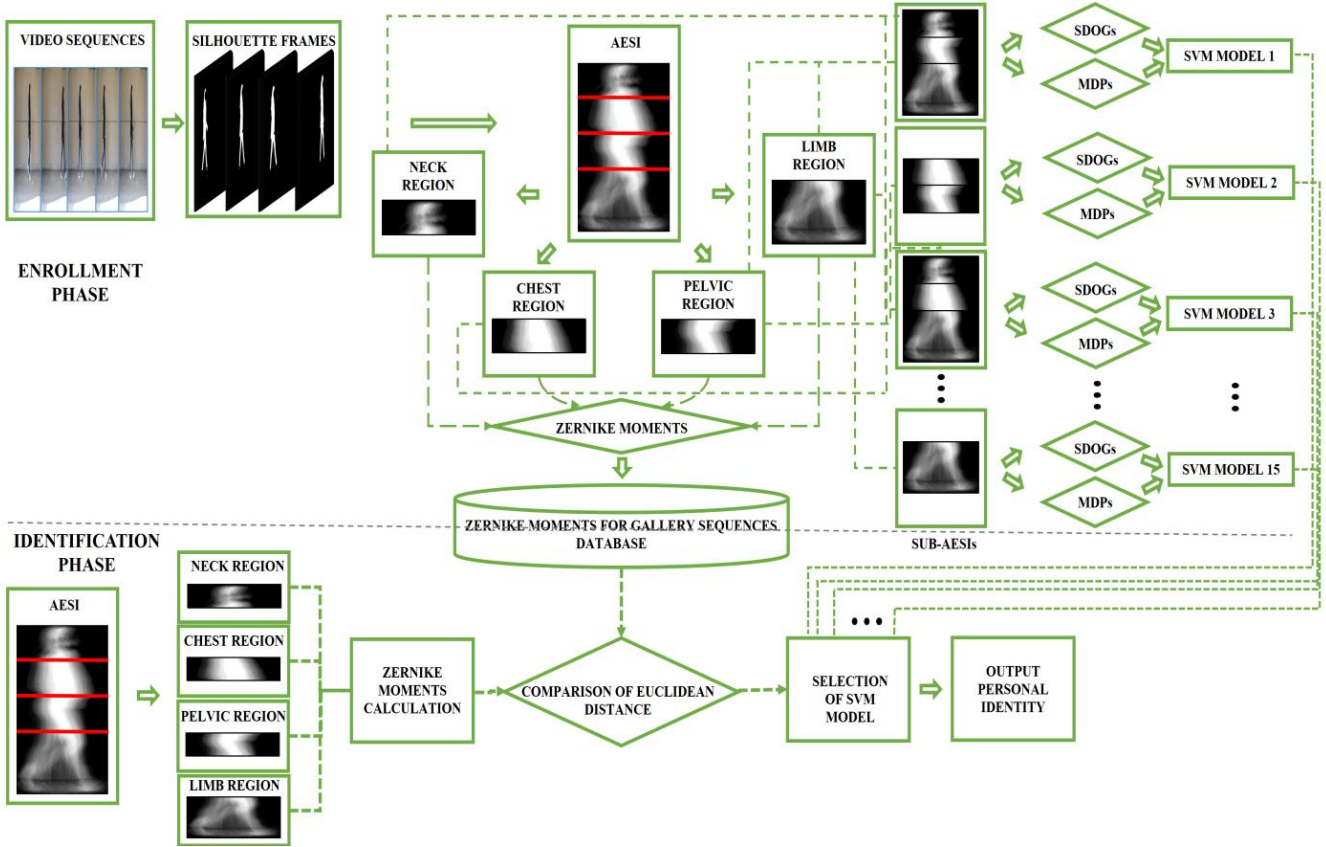

Fig.1. Work flow diagram of the proposed framework

Liu et al. *[28]* modelled human gaits using population Hidden Markov Model (pHMM) and calculated distance between sums of normalized gait stances to identify individual identity. Wang et al. *[7]* introduced a novel template named Chrono-Gait Image (CGI). CGI encodes variation in averaged images into multiple channels thereby effectively preserving the temporal information of gait motion. One of the principal challenge in gait recognition is dealing with intra-class variations spawning from clothing variations and carrying conditions. Several techniques have been studied to reduce adverse effects of these variations. Bashir et al. *[9]* proposed Gait Entropy image (GEnI). GEnI calculates entropy for each pixel in GEI and then segments high temporal regions from regions of low temporality. The regions of high temporality are argued to be devoid of covariate conditions and thus are more relevant to gait recognition. Bashir et al. *[8]* used optical flow fields to capture motion intensity and motion direction. Dupuis et al. *[29]* selected relevant features from GEI by using Random Forest feature rank algorithm. Jeevan et al. *[30]* presented novel temporal template named Gait Pal and Pal Entropy (GPPE) Image. Kusakunniran *[31], [32]* proposed Space Time interest points (STIPs) descriptor to extract gait features directly from raw video. Rokanujjaman et al. *[33]* introduced a novel discrete fourier transform based entropy representation of gait named EnDFT. Lam et al. *[17]* proposed information set theory which involve features derived from Hanman-Anirban entropy function. GII have two features- energy feature (GII-EF) and sigmoid feature (GII-SF). Guan et al. *[34]* proposed a classifier ensemble method based on the random subspace method (RSM) and majority voting (MV). Very recently, Choudhury and Tjahjadi *[35]* introduced averaged gait key-phase image (AGKI) which use rotation forest ensemble learning technique to distinguish individuals and recognize intra-class diversity, leading to good results even in presence of covariates.

III. PROPOSED COVARIATE CONSCIOUS FRAMEWORK

The covariate conscious gait recognition approach can be understood in two phases: Enrollment phase and Identification phase as outlined in Fig. 1.

In Enrollment phase, AESI is formed for each of the gallery gait sequence. The formed AESI is segmented into 4 parts according to human body geometry namely neck region, chest region, pelvic region and limb region. Each region is termed as part-AESI. Zernike moments are then calculated for each of the part-AESI and added in '*Zernike moments for gallery sequences*' database. In parallel, sub-AESIs are formed by arranging part-AESIs into all possible combinations resulting in a total of $2^4 - 1 = 15$ sub-AESIs. Features are then extracted from each of the sub-AESI using SDOGs and MDPs methods. Finally, a linear SVM classifier is trained on the obtained feature set.

In Identification phase, AESI is formed from the probe gait sequence. Similar to enrollment phase, part-AESIs are constructed and Zernike moments are computed for each of the part-AESI. The presence of covariates is then decided by comparison of the computed moments of a part-AESI with the moments stored in '*Zernike moments for gallery sequences*' database for the corresponding part. As decided by the comparison, parts devoid of any covariates are used to form a sub-AESI which is later classified using the corresponding SVM classifier trained during the enrollment phase.

*A. Construction of AESI*

AESI [14] preserves shape and temporal information that is representative of gait motion in a convenient 2D template. AESI use binary silhouettes as basis and its formation is a two-step process. The first step involves extraction of human silhouettes from image frames and calculation of gait period. The second step involves summation and normalization of the obtained silhouettes.

The extraction of binary silhouettes is performed using background subtraction technique. Background subtraction is a challenging task owing to the highly dynamic nature of the background, at the same time, quality of the extracted silhouettes is critical for the performance of the whole gait recognition algorithm. This makes effective modelling of background imperative. This work employs textual feature based background subtraction approach proposed by Chua et al. [36] for foreground segmentation. The approach reckons entropy as an important texture information parameter, which is defined as:

$$\mathcal{E} = \sum_i \sum_j \mathcal{F}(i,j) \log \mathcal{F}(i,j) \quad (1)$$

where $\mathcal{F}(i,j) = \frac{\mathcal{M}(i,j)}{\sum_{i,j} \mathcal{M}(i,j)}$, is the probability density function and, $i$ and $j$ are indices to co-occurrence matrix $\mathcal{M}$. Once the binary images are formed, the silhouettes are obtained by scanning the image row-wise and column-wise. The extreme white pixels are then determined to get the desired bounding box.

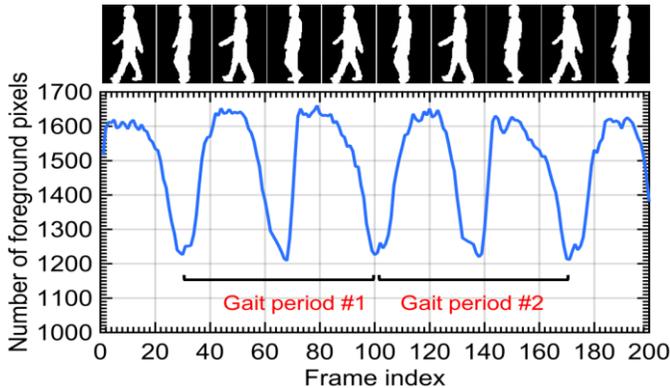

Fig.2. Determination of gait period from variation of foreground pixels

One key step in formation of AESI for gait sequence is determination of gait period. Gait period is defined as the number of frames over which two consecutive walking cycles span. One walking cycle is completed when a person moves from mid-stance position (position where legs are closest) to double-support position (legs are farthest apart) and then back to mid-stance position. The gait period can be computed simply by counting the number of foreground pixels in the lower half [21]. At the mid-stance position the number of foreground pixels attains a local minima, while at double-support position the number attains a local maxima. Therefore, the analysis as shown in Fig. 2 of these local minima and maxima can give the estimate of the gait period.

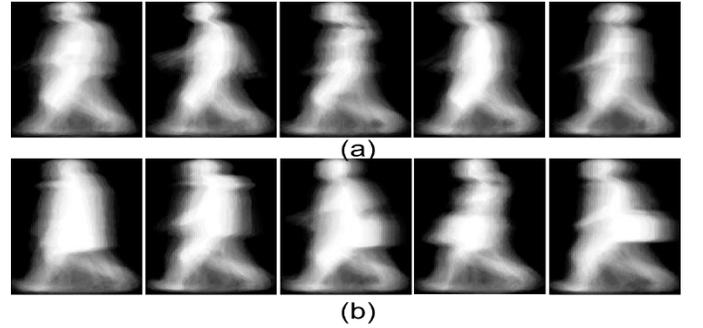

Fig.3. Formed AESIs (a) Normal Walking (b) Clothing and carrying

Once the gait period is determined, the binary silhouettes over that period are aligned using reference point and then fused into a single template on incremental basis [37]. One of the key strengths of AESI is the degree of compression it offers, an entire gait sequence, typically 40-70 silhouettes, is compressed into a single template. AESI fails to maintain the chronology thus loosing on some of the temporal information, however, it has adequate shape information thus making it apposite for the task of gait recognition. AESI is defined as:

$$AESI(x,y) = 1/n \sum_{\tau=1}^{n} |\mathfrak{S}(x,y)_\tau|^2 \quad (2)$$

where '$n$' is the number of silhouettes in a gait period and '$\mathfrak{S}(x,y)_\tau$' is the binary silhouette at time instant '$\tau$'. The salient characteristic of AESI is that the degree of temporality in a region is manifested as intensity variation in AESI. Examples of generated AESIs for normal walking and covariate conditions is shown in Fig. 3.

*B. Covariate detection using ZMIs*

The presence of covariates can significantly alter the shape of human silhouettes. While even though this is one of the prime reason for drop in performance of gait recognition systems, this information could actually be leveraged in order to filter out the regions of the silhouettes affected by such distortion. In this paper, we use Zernike moment based shape

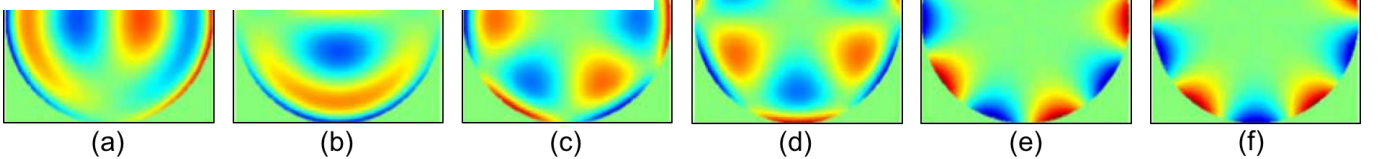

Fig.4. Plot of Zernike Basis function $\mathcal{V}_{nm}(\rho,\theta)$ for order $n=5$  (a) Re($\mathcal{V}_{5,1}$) (b) Im($\mathcal{V}_{5,1}$) (c) Re($\mathcal{V}_{5,3}$) (d) Im($\mathcal{V}_{5,3}$) (e) Re($\mathcal{V}_{5,5}$) (f) Im($\mathcal{V}_{5,5}$)

descriptors for the purpose. The incentive to using Zernike moments is that they are invariant to position, size and orientation. Teh et al. [38] demonstrated the efficacy of Zernike moments as shape descriptors among other many moment-based descriptors. Zernike polynomials are a set of complex polynomials forming an orthogonal basis, which are defined in the interior of unit circle $x^2 + y^2 \leq 1$. Zernike basis function $\mathcal{V}_{nm}(\rho, \theta)$ of order $n$ and repetition $m$ is defined in polar coordinates as:

$$\mathcal{V}_{nm}(x, y) = \mathcal{V}_{nm}(\rho, \theta) = \mathcal{R}_{nm}(\rho)e^{jm\theta} \quad \text{for } \rho \leq 1 \quad (3)$$

where '$\mathcal{R}_{nm}(\rho)$' is a radial polynomial defined as:

$$\mathcal{R}_{nm}(\rho) = \sum_{s=0}^{\frac{n-|m|}{2}} (-1)^s \frac{(n-s)!}{s!\left(\frac{n+|m|}{2}-s\right)!\left(\frac{n-|m|}{2}-s\right)!} \rho^{n-2s} \quad (4)$$

Here, $n$ is a non-negative number and $m$ is integer satisfying conditions: $n - |m|$ is even and $|m| \leq n$. The set of basis functions $\{\mathcal{V}_{nm}(\rho, \theta)\}$ is orthogonal, i.e.

$$\int_0^{2\pi} \int_0^1 \mathcal{V}^*_{nm}(\rho, \theta) \mathcal{V}_{pq}(\rho, \theta) \rho d\rho d\theta = \frac{\pi}{n+1} \delta_{np} \delta_{mq} \quad (5)$$

where $\delta_{xy} = \begin{cases} 1, & x = y \\ 0, & otherwise \end{cases}$.

The 2-dimensional Zernike moments for continuous image function $\psi(\rho, \theta)$ are defined as:

$$Z_{nm} = \frac{n+1}{\pi} \int_0^{2\pi} \int_0^1 \psi(\rho, \theta) \mathcal{V}^*_{nm}(\rho, \theta) \rho d\rho d\theta \quad (6)$$

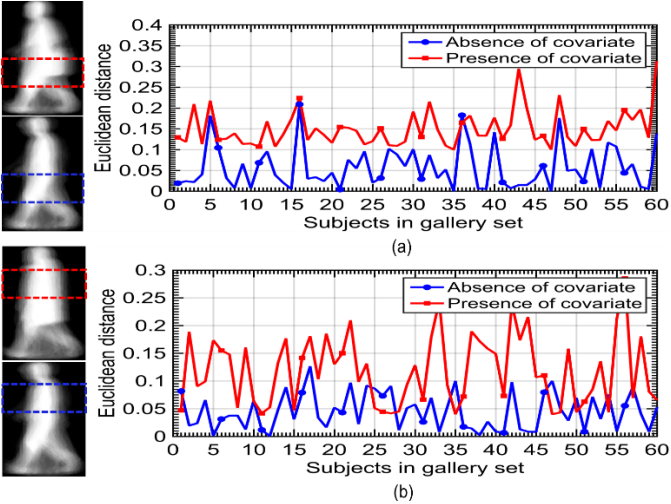

Fig.5. Variation of Euclidean distance between Zernike moments in presence and absence of covariates for two instances (a) and (b). From figure it can be seen that the mean of Euclidean distance does serve as a reliable metric to detect the presence of covariates

$$= \frac{n+1}{\pi} \int_0^{2\pi} e^{-jm\theta} \int_0^1 \psi(\rho, \theta) \mathcal{R}_{nm}(\rho) \rho d\rho d\theta$$

The 2-dimensional Zernike moments for discrete digital image function $\psi(x, y)$ are defined as:

$$Z_{nm} = \frac{n+1}{\pi} \sum_x \sum_y \psi(x, y) \mathcal{V}^*_{nm}(x, y), x^2 + y^2 \leq 1 \quad (7)$$

As stated before, the AESIs of both the gallery and probe gait sequences are partitioned into 4 segments called part-AESIs. This is done in order to restrict the extent of discarded region of AESI to a minimum, and thereby preserving viable information as much as possible. Since the AESIs are size normalized, the same segmentation scheme is followed for all the images. An AESI of height H is segmented based on human body anatomy [39] and scheme is given in Table I.

TABLE I
AESI SEGMENTATION

| Body Part | Height |
| --- | --- |
| Neck region | 1H - 0.80H |
| Chest region | 0.80H - 0.55H |
| Pelvic region | 0.55H - 0.30H |
| Limb region | 0.30H - 0H |

Once the partition is done, Zernike moment invariants are calculated for each part-AESI formed for the gallery sequences, and are stored in '*Zernike moments for gallery sequences*' database. To detect the presence of covariate in incoming part-AESI $P_i$ ($i = 1, 2 \ldots 4$) of probe gait sequence, we compute mean Euclidean distance $D_{Pi}$ between its Zernike moments and the moments stored in '*Zernike moments for gallery sequences*' database for the corresponding part. In our empirical testing we found Zernike moments of order $n = 5$ and repetition $m = 1, 3$ and $5$ to be efficient for the comparison. Fig. 4 shows plots of real and imaginary part of these moments. $D_{Pi}$ is defined as:

$$D_{Pi} = \frac{1}{N} \sum_{k=1}^{N} \sqrt{\sum_{\substack{m=1,3,5 \\ n=5}} |Z_{nm}^{Gki} - Z_{nm}^{Pi}|^2} \quad (8)$$

where $N$ is the total number of gallery sequences and $G_{Ki}$ is $i^{th}$ part –AESI of $K^{th}$ gallery sequence.

A part-AESI is declared to infected with covariate if:

$$D_{Pi} \geq \mu_i + 3\sigma_i \quad (9)$$

where:

$$\mu_i = \frac{1}{N^2} \sum_{j=1}^{N} \sum_{\substack{k=1 \\ k \neq j}}^{N} \sqrt{\sum_{\substack{m=1,3,5 \\ n=5}} |Z_{nm}^{Gji} - Z_{nm}^{Gki}|^2} \quad (10)$$

$$\sigma_i = \frac{1}{N} \sum_{j=1}^{N} \left( \sum_{\substack{k=1 \\ k \neq j}}^{N} \sqrt{\sum_{\substack{m=1,3,5 \\ n=5}} |Z_{nm}^{Gji} - Z_{nm}^{Gki}|^2} - \mu_i \right) \quad (11)$$

It should be noted that computation of $\mu_i$ and $\sigma_i$ is an offline process, and needs to be executed only once, when the database is constructed. This work considers the prevalent scenario where gallery is formed of normal walking sequences and probe sequences involve presence of covariates. However, the same proposed algorithm can be used in the hypothetical opposite scenario as well, since covariate detection works only on relative distance between probe and gallery sequences.

Fig. 5 demonstrates variation of Euclidean distance for infected and non-infected part-AESIs for two instances. It can be discerned from the figure that the mean Euclidean distance does serve as a suitable metric for the detection of covariates.

The added advantage of using ZNIs on AESI rather than on binary silhouettes themselves is AESI's immunity to noise. Since, AESI is formed by fusing multiple silhouettes; the effect of noise in any one silhouette becomes negligible and hence does not perniciously influence the performance of Zernike shape descriptors.

*C. Computation of SDOGs shape features*

SDOG [14] extracts textural information from images by binning the intensity of image gradients according to their orientation in the image. These characteristics are computed at various decomposition levels (sub-regions) of the AESI, and hence encapsulate both coarse and fine details. The SDOGs computation for $K$ orientation bins at decomposition level $v$ results in a feature vector of length $K \sum_v 4^v$. In this work, we apportion gradient orientations into 9 bins each spanning 20
5

degrees and a total of 3 decomposition levels ($v = 0, 1, 2$) are employed. The process to compute SDOG is delineated in Algorithm1 and Fig. 6.

The reason why SDOG particularly complements AESI as a feature extraction method is its fixation to image gradients. Since, AESI manifests both temporal and shape information of gait sequence as intensity variation, it's the AESI's gradients which possess the needed information for optimal recognition performance.

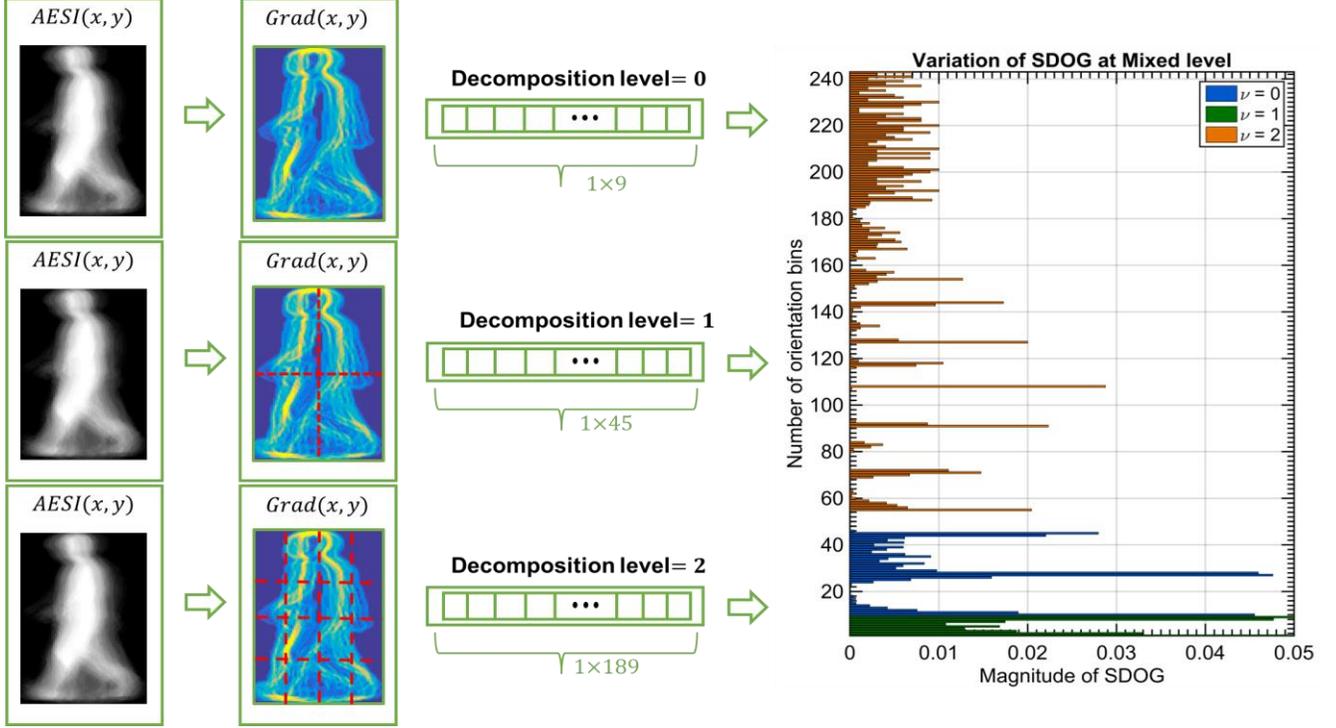

Fig. 6. Representation of decomposition levels and their spatial distribution, involved in calculation of final SDOG feature vector

**Algorithm 1: Computation of SDOGs**

**Input.** $AESI(x, y)$ of gait sequence computed using Eq.2.
**Step1.** Decompose $AESI(x, y)$ into sub-regions.
**Step2.** Determine the magnitude and orientation at each pixel of gradient image $Grad(x, y)$ at multiple decomposition levels (v) using $\mathcal{H}(x, y) = \sqrt{Grad_x(x,y)^2 + Grad_y(x,y)^2}$ and $\emptyset(x, y) = \tan^{-1}\left[\frac{Grad_y(x,y)}{Grad_x(x,y)}\right]$.
**Step3.** Apportion the computed magnitude of gradients at different levels (v) into $K$- orientation bins.
**Step4.** Aggregate all the sub-regions at level (v) $f_{SDOGs} = K\sum_v 4^v$
**Step 5.** Concatenate all the levels (0, 1, 2) of SDOGs to form a final SDOGs feature vector defined as: $f_{SDOGs} = K\left(\underbrace{\sum_v 4^v}_{\text{level v=0}} ; \underbrace{\sum_v 4^v}_{\text{level v=1}} ; \underbrace{\sum_v 4^v}_{\text{level v=2}}\right)$.
**Output:** The histogram of SDOGs feature vector of dimension $1 \times 189$.

### D. MDPs calculation

Human gait motion is attributed more to horizontal motion than vertical motion [16]. To exploit this, we introduce MDPs features, and append these with SDOGs to form the final feature set. MDPs treat each row of AESI as an individual feature element, and compute the mean of pixel intensity in individual rows. To keep the dimension of the resulting feature set low, only the rows of AESI which belong to high temporal regions, are included in the computation. High temporal rows in AESI can be identified from the variance of intensity values in that particular row.

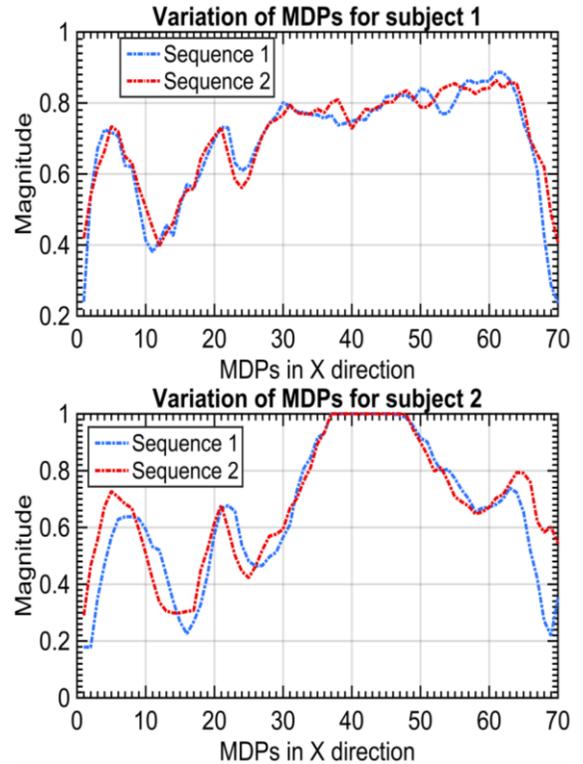

Fig.7. Variation of MDPs for two subjects. The figure illustrates the intra-class similarity and inter-class distinguishing ability possessed by MDPs.

For AESI of dimension $m \times n$, $MDP(i)$ for row vector $i$ is defined as:

$$MDP(i) = \frac{1}{n}\sum_{k=1}^{n} AESI(i,k) \qquad (12)$$

In this work, only $m/2$ MDPs corresponding to most intensity variable rows, are calculated for a $m \times n$ AESI.

It should be noted that since the number of rows vary for different part-AESIs, the length of resulting feature set varies as well. Fig. 7 demonstrates the variation of MDPs for two subjects with two sequences each. It can be studied from the figure that MDPs possess remarkable inter-class distinguishing ability while at the same time have the needed intra-class similarity.

## IV. EXPERIMENTAL RESULTS

In order to test the performance of proposed framework, an experiment is conducted on three publicly available human gait dataset sets. These datasets are CASIA Dataset B [20], OU-ISIR Treadmill Dataset B [18] and the USF Human-ID gait challenge Dataset [21]. Performance of the algorithm is measured by calculating the correct class recognition rate (CCR) in percentage using support vector machine (SVM) as linear kernel in leave-one-out cross validation scheme. The CCR is determined using Eq. 14 and defined as:

$$CCR = \frac{TP+TN}{TP+TN+FP+FN} \times 100 \; (In \; Percentage) \qquad (14)$$

where TP, TN, FP, and FN are true positive, true negative, false positive, and false negative respectively. To evaluate the effectiveness of the proposed algorithm, the highest CCR achieved through the proposed algorithm is compared with the others similar state-of-the-art.

### A. CASIA Dataset B

CASIA dataset B [20] is a multi-view dataset comprising of 124 subjects. The gait motion is captured from 11 cameras placed 18 degrees apart. There are 10 sequences of each subject which include 6 normal walking sequences (CAS-A), and 2 each with carrying conditions (CAS-B) and clothing

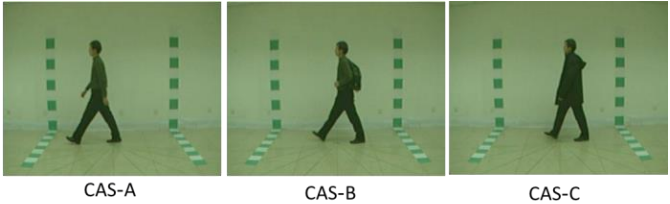

Fig.8. Probe sequences in CASIA B dataset

variation (CAS-C).

For evaluation, only sequences under 90 degrees' view angle have been considered. The first 4 sequences of CAS-A are used as gallery sequences while the remaining 2 along with sequences of CAS-B and CAS-C form the probe set. The AESIs of all the gait sequences are size normalized to $136 \times 72$ pixels.

The highest CCR achieved in each probe of the dataset is compared with the others similar state-of-the-art and as presented in Table II and Fig. 9.

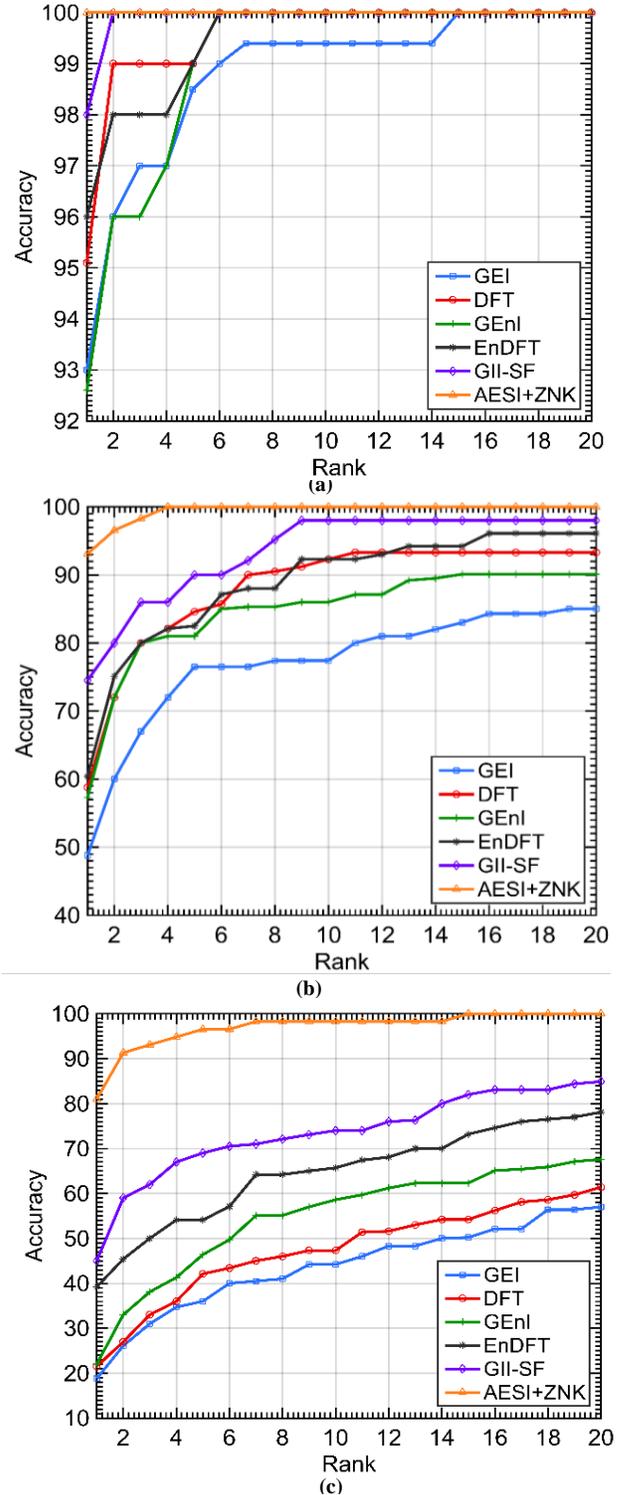

Fig. 9. Depiction of accuracy vs Rank curve for different approaches on three probe sequences of dataset (a) CAS-A, (b) CAS-B (c) CAS-C

Table II, compares the performance of the proposed method with various existing algorithms in terms of average CCR for three different probe of CASIA dataset. From Table II, it can be observed that the proposed algorithm performs admirably

well in presence of covariates (i.e. CAS-B and CAS-C), where it achieves 93.1% accuracy for sequences having carrying conditions (CAS-B). This is 9.5% better than the existing state-of-the-art, Flow Field which achieved recognition accuracy of 83.6%. For sequences having clothing variations (CAS-C), difference is even more drastic. Our method achieved accuracy of 81.3%. This is 17.6% better than the existing Random forest approach, which achieved recognition accuracy of 63.7%. Fig. 9 show Cumulative Match Characteristic (CMC) curve for all three probe sequences, and it can be considered an exhaustive comparison of accuracy because it is achieved through the variation of Rank. In all these cases the CMC curves shows better accuracy. Hence, it can be said the proposed algorithms perform well in comparison with many such existing approaches.

TABLE II
COMPARISON OF CCR WITH THE SIMILAR STATE-OF-THE-ART ON CASIA DATASET B

| Method | CAS-A | CAS-B | CAS-C | Average |
|---|---|---|---|---|
| GEI [6] | 93.1 | 48.8 | 18.8 | 53.56 |
| GEnI [9] | 92.6 | 57.3 | 22.0 | 57.30 |
| DFT [40] | 95.1 | 58.8 | 21.5 | 58.46 |
| EnDFT [33] | 96.0 | 60.4 | 39.2 | 65.20 |
| CGI [7] | **100** | 64.2 | 43.6 | 69.26 |
| Flow Field [29] | 97.5 | 83.6 | 48.8 | 76.63 |
| Random Forest [29] | 98.8 | 73.8 | 63.7 | 78.76 |
| GPPE [30] | 93.6 | 56.1 | 22.4 | 57.36 |
| STIP [32] | 95.4 | 60.9 | 52 | 69.43 |
| GII-SF [19] | 98.0 | 74.5 | 45 | 72.5 |
| **AESI+ZNK** | **100** | **93.1** | **81.3** | **91.47** |

### B. OU-ISIR Treadmill Dataset B

OU-ISIR Treadmill B [18] dataset comprises of large clothing variations. The dataset consists of 68 subjects with up to 32 possible variations in clothing from 15 different kinds of clothes such as half shirt, full shirt, baggy pants, hat, cap etc. Fig. 10 shows different gait sequences in the dataset. Table III show clothing combinations for each of the gait sequence.

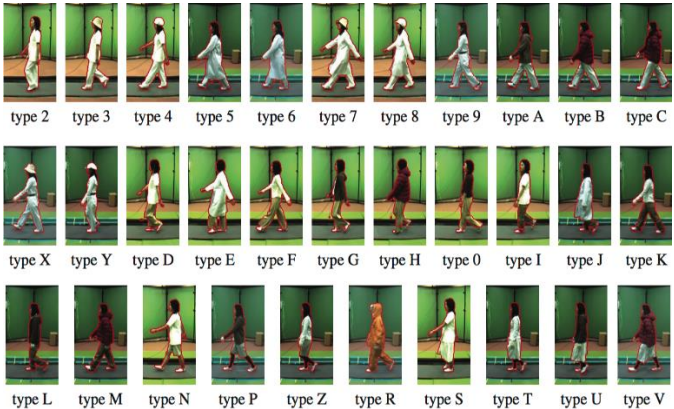

Fig. 10. Gait sequences in OU-ISIR Treadmill Dataset B

The dataset is partitioned into 3 segments- Gallery set, Training set and Probe set. The gallery set comprises of 48 subjects with one standard combination of clothing (Sequence type 9). The training set consists of 446 sequences of 20 subjects with all possible clothing combinations. The set is provided for tuning of model parameters. Since, its unlikely in real world to know the type of covariate variations beforehand, the training set is not used in this evaluation in any manner. The probe set comprises of 856 sequences of the same 48 subjects as in gallery set with all possible clothing combinations, except type 9. The OU-ISIR dataset itself provide extracted silhouettes size normalized to $128 \times 88$ pixels, therefore the formed AESIs are of the same size ($128 \times 88$).

TABLE III
CLOTHING COMBINATIONS INCLUDED IN THE OU-ISIR B DATASET. KEYS FOR CLOTHES: RP-REGULAR PANTS; BP-BAGGY PANTS; SP-SHORT PANTS; SK-SKIRT; CP: CASUAL PANTS; HS-HALF SHIRT; FS- FULL SHIRT; LC-LONG COAT; PK-PARKER; DJ-DOWN JACKET; CW-C

| Type | $S_1$ | $S_2$ | $S_3$ | Type | $S_1$ | $S_2$ | Type | $S_1$ | $S_2$ |
|---|---|---|---|---|---|---|---|---|---|
| 3 | RP | HS | Ht | 0 | CP | CW | F | CP | FS |
| 4 | RP | HS | Cs | 2 | RP | HS | G | CP | Pk |
| 6 | RP | LC | Mf | 5 | RP | LC | H | CP | DJ |
| 7 | RP | LC | Ht | 9 | RP | FS | I | BP | HS |
| 8 | RP | LC | Cs | A | RP | Pk | J | BP | LC |
| C | RP | DJ | Mf | B | RP | DJ | K | BP | FS |
| X | RP | FS | Ht | D | CP | HS | L | BP | Pk |
| Y | RP | FS | Cs | E | CP | LC | M | BP | DJ |
| N | SP | HS | - | P | SP | Pk | R | RC | - |
| S | Sk | HS | - | T | Sk | FS | U | Sk | PK |
| V | Sk | DJ | - | Z | SP | FS | - | - | - |

The Table IV compares the rank 1 and rank 5 performance of the proposed method with the similar state of the art. Fig.11 shows the comparison of CMC on similar state-of-the-art methods and higher accuracy can be comprehend by the proposed approach AESI+ZNK.

The results are depicted using CMC curve in Fig. 12. shows the CMC curve for all the probe sequences in the OU-ISIR Treadmill B dataset achieved by our algorithm.

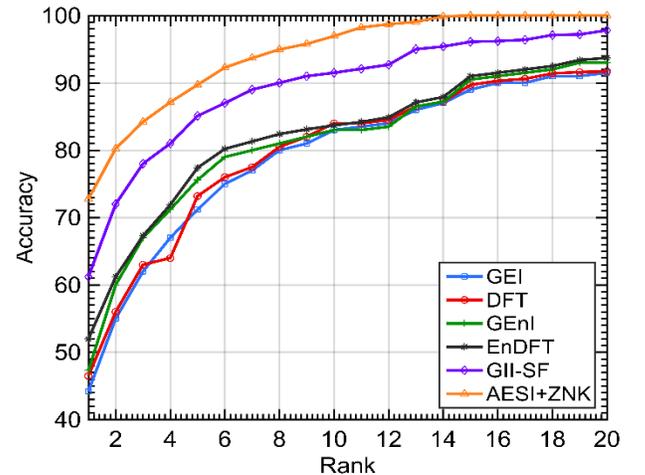

Fig. 11 Comparison using CMC on OU-ISIR Treadmill B dataset

TABLE IV
COMPARISON OF CCR WITH THE SIMILAR STATE-OF THE-ART ON OU-ISIR TREADMILL DATASET B

| Method | Rank 1 CCR | Rank 5 CCR |
|---|---|---|
| GEI [5] | 44.2 | 71.2 |
| GEnI [9] | 47.3 | 75.6 |
| DFT [37] | 46.6 | 73.2 |
| EnDFT [32] | 52.0 | 77.4 |
| GII-SF [18] | 61.2 | 85.1 |
| **AESI+ZNK** | **72.7** | **89.7** |

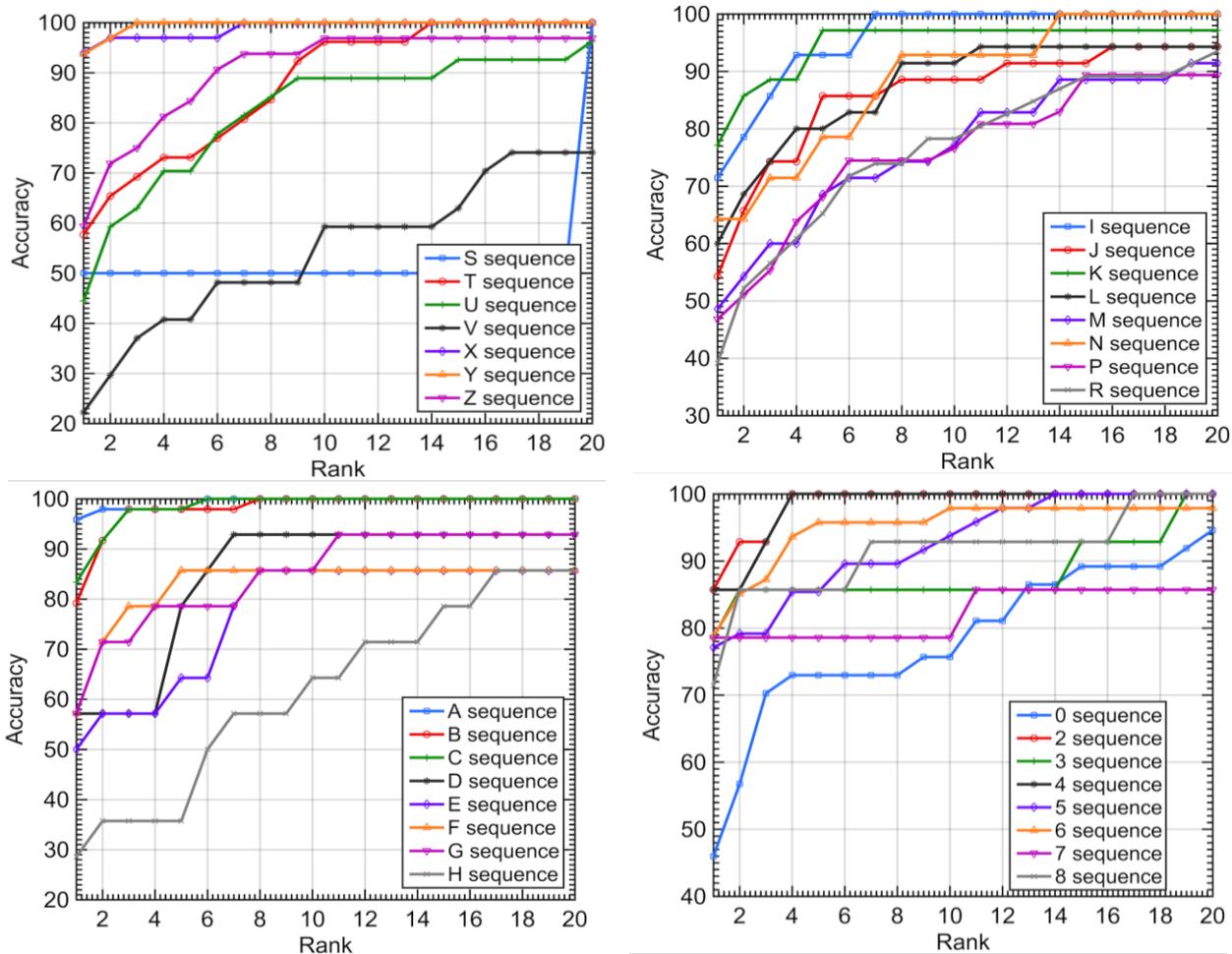

Fig. 12. Comparison of performance of proposed algorithm for individual sequences of OU-ISIR Treadmill B Dataset using CMC

It is clear from Table IV that the proposed approach outperforms existing methods in robustness to distortion by presence of covariates. The recognition accuracy achieved by the proposed algorithm is 72.7%, which is 11.5% better than the existing state-of-the-art, GII-SF that achieved 61.2% recognition accuracy.

From Fig. 12 it can be seen that the recognition accuracy achieved is highest for probe sequences A, X and Y. This is understandable since these sequences have less drastic variations in clothing compared to gallery set. On the contrary, the proposed algorithm fared worst for sequences V and H. This can be attributed to minimal commonality between these sequences and the gallery set.

### C. USF Human-ID Gait Challenge Dataset

Finally, the performance of the proposed work is evaluated on USF Human-ID dataset [20]. The dataset is shot in outdoor environment (Fig. 13) and comprises of five variations- concrete or grass surface, presence or absence of briefcase, shoe variation (Type A or Type B), right or left view and variation in elapse time.

The USF Human-ID gait challenge dataset contains a total of 1870 gait sequences of 122 human subjects. The dataset provides a predetermined experimental setup and is segmented into 13 subsets out of which 1 forms the gallery set and the remaining 12 form the probe sets (A-L). As in the case of OU-ISIR dataset, the USF Human-ID dataset also provides size-normalized silhouettes of $128 \times 88$ pixels. Therefore, the formed AESIs are of the same size i.e. $128 \times 88$ pixels.

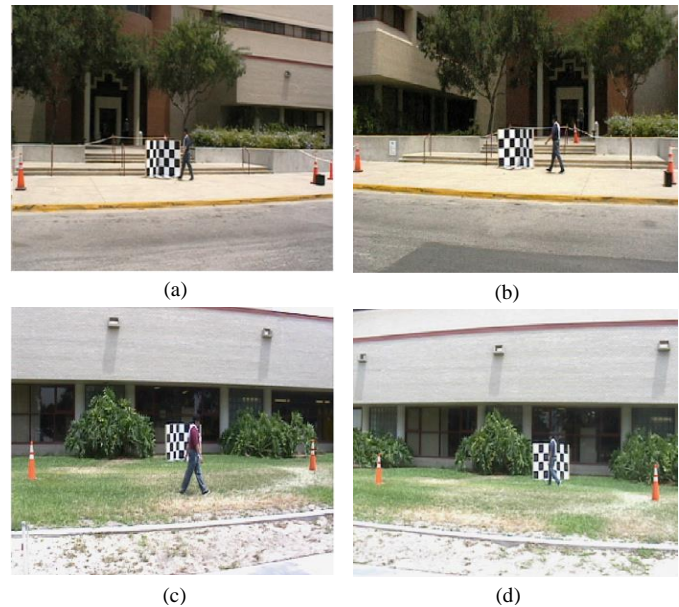

Fig.13. Sample images of USF Human-ID gait challenge dataset: (a, b) left and right of camera view on concrete surface,(c, d) left and right camera view on grass surface

Table V informs about the covariate variations in all the probe sets (A-L) along with the number of sequences in each set. The table also demonstrates the achieved overall accuracy of the proposed algorithm and similar state-of-the-art for individual probe sets.

TABLE V
COMPARISON OF CCR WITH THE SIMILAR STATE-OF THE-ART ON USF HUMAN-ID GAIT DATASET AT RANK 1 AND RANK 5. KEY FOR COVARIATES -V - VIEW; H - SHOE; S - SURFACE; B - BRIEFCASE; T - TIME; AND C - CLOTHES

| Probe Set | A | B | C | D | E | F | G | H | I | J | K | L | Avg. |
|---|---|---|---|---|---|---|---|---|---|---|---|---|---|
| Probe Size | 122 | 54 | 54 | 121 | 60 | 121 | 60 | 120 | 60 | 120 | 33 | 33 | - |
| Covariates | V | H | VH | S | SH | SV | SHV | B | BH | BV | THC | STHC | - |
| Rank 1 CCR | | | | | | | | | | | | | |
| GEI [5] | 90 | 91 | 81 | 56 | 64 | 25 | 36 | 64 | 60 | 60 | 6 | 15 | 57.66 |
| CGI [7] | 91 | 93 | 78 | 51 | 53 | 35 | 38 | 84 | 78 | 64 | 3 | 9 | 61.69 |
| GTDA [10] | 91 | 93 | 86 | 32 | 47 | 21 | 32 | 95 | 90 | 68 | 16 | 19 | 60.58 |
| GPDF [38] | 95 | 93 | 89 | 62 | 62 | **39** | 38 | 94 | 91 | 78 | 21 | 21 | 70.07 |
| VI-MGR [34] | 95 | 96 | 86 | 54 | 57 | 34 | 36 | 91 | 90 | 78 | 31 | 28 | 68.13 |
| AGKI [21] | 96 | 96 | 90 | 62 | 63 | 37 | **39** | 94 | **93** | 80 | **41** | **32** | 71.74 |
| **AESI+ZNK** | **97** | 96 | **93** | **68** | 64 | 34 | 37 | **96** | 92 | **86** | 27 | 24 | **72.53** |
| Rank 5 CCR | | | | | | | | | | | | | |
| GEI | 94 | 94 | 93 | 78 | 81 | 56 | 53 | 90 | 83 | 82 | 27 | 21 | 76.23 |
| CGI | 97 | 96 | 94 | 77 | 77 | 56 | 58 | 98 | 97 | 86 | 27 | 24 | 79.12 |
| GTDA | 98 | 99 | 97 | 68 | 68 | 50 | 56 | 95 | 99 | 84 | 40 | 40 | 77.58 |
| GPDF | 99 | 94 | 96 | **89** | **91** | 64 | 64 | 99 | 98 | 92 | 39 | 45 | 85.31 |
| VI-MGR | 100 | 98 | 96 | 80 | 79 | 66 | 65 | 97 | 95 | 89 | 50 | 48 | 83.75 |
| AGKI | 100 | 98 | 97 | 88 | 85 | 68 | **68** | 98 | 95 | 91 | **57** | **54** | 84.46 |
| **AESI+ZNK** | 100 | **100** | **98** | 81 | 83 | 68 | 60 | **99** | **98** | **94** | 42 | 36 | **84.67** |

It can be studied that the proposed algorithm achieved an overall CCR of 72.53% for Rank 1 evaluation. This is highest among all the compared methods. For Rank 5 evaluation, the proposed method achieved CCR of 84.67%, which again is highest among all the existing methods. This proves the efficacy of our approach. The achieved CCR is particularly high for probe sets A, C, D, H and J. However, the approach doesn't offer improved performance for probe sets K and L over the existing state-of-the-art. This shows that while the use of Zernike moments can effectively handle and iron out covariates affecting the shape information, it understandably doesn't offer much immunity from temporal covariates.

## V. CONCLUSION AND FUTURE WORK

In this paper, a covariate cognizant approach to gait recognition has been presented. A single spatio-temporal template called AESI is formed to represent the gait motion. After which, Zernike moment based shape descriptors are used to detect the presence of covariates. An effective segmentation technique is used which discards the parts of AESI deemed infected by covariates. For feature extraction, SDOGs and MDPs methods are employed. SDOGs method provides a reliable and compact feature set by exploiting gradient information in the AESI. MDPs is a new feature extraction method, adapted specifically for gait motion and extracts information about the horizontal movement, which characterize the gait sequence. The efficacy of the presented approach is evaluated on three publicly available datasets. On all the three datasets our approach showed promising results and outperformed the existing state-of-the-art.

For future studies, an extension of the proposed approach, which is multi-view invariant, can be developed. We will also explore how to improve the representation of temporal information for gait sequences. In addition, we will consider employing the presented approach to the domain of medical prognosis. Since many ailments causes substantial change in walking pattern of an individual, it would be interesting to explore if the technique could help in diagnosis of such ailments.

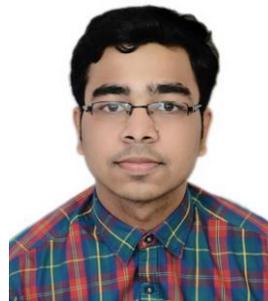

**Himanshu Aggarwal** received his B.Tech. degree in Electronics and Communication Engineering from Delhi Technological University in 2016. Presently, he is working as Associate software engineer in Computer Vison department of Qualcomm Incorporated, Hyderabad, India. His current interest includes human activity recognition, egocentric visual odometry and algorithmic design.

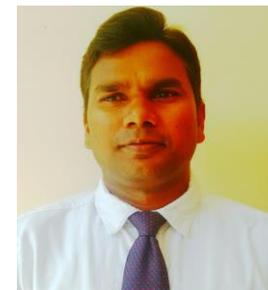

**Dinesh Kumar Vishwakarma** received his B.Tech. from Dr RML Avadh University, Faizabad, Uttar Pradesh, India, in 2002, M.Tech. from Motilal Nehru National Institute of Technology, Allahabad, Uttar Pradesh, India in 2005, and Ph.D. in Computer Vision from Delhi Technological University, Delhi, India in 2016. Presently, he is working as Assistant Professor, in the Department of Electronics and Communication Engineering, Delhi Technological University, Delhi, India-110042. His research interest includes human action and activity recognition, hand gesture recognition and machine learning.